\documentclass[10pt,twocolumn,letterpaper]{article}

\usepackage{cvpr}
\usepackage{times}
\usepackage{epsfig}
\usepackage{graphicx}
\usepackage{amsmath}
\usepackage{amssymb}
\usepackage{graphicx}
\usepackage{amsmath,amssymb,enumitem} 

\usepackage[pagebackref=true,breaklinks=true,letterpaper=true,colorlinks,bookmarks=false]{hyperref}

 \cvprfinalcopy 

\def\cvprPaperID{6203} 

\ifcvprfinal\fi
\begin{document}
	
	\title{Deep Transfer Learning for Multiple Class Novelty Detection}
	
\author{Pramuditha Perera and Vishal M. Patel\\
	Department of Electrical and Computer Engineering,\\
	Johns Hopkins University, Baltimore, MD 21218, USA\\
	{\tt\small  pperera3@jhu.edu, vpatel36@rutgers.edu}
		\thanks{This work was supported by the NSF
		grant 1801435.}
}
	
	\maketitle
	

	\begin{abstract}
		
We propose a transfer learning-based solution for the problem of multiple class novelty detection.  In particular, we propose an  end-to-end deep-learning based approach in which we investigate how the knowledge contained in an external, out-of-distributional dataset can be used to improve the performance of a deep network for visual novelty detection.  Our solution differs from the standard deep classification networks on two accounts. First, we use a novel loss function, \textit{membership loss}, in addition to the classical cross-entropy loss for training networks. Secondly, we use the knowledge from the external dataset more effectively to learn \textit{globally negative filters},  filters that respond to generic objects outside the known class set. We show that thresholding the maximal activation of the proposed network can be used to identify novel objects effectively. Extensive experiments on four publicly available novelty detection datasets show that the proposed method achieves significant improvements over the state-of-the-art methods.

	\end{abstract}

	\section{Introduction}
	In recent years, intelligent systems powered by artificial intelligence and computer vision that perform visual recognition  have gained much attention \cite{he15deepresidual},\cite{NIPS2012_ALEX},\cite{VGG}.   These systems observe instances and labels of known object classes during training and learn association patterns that can be used during inference. A practical visual recognition system should first determine whether an observed instance is from a known class. If it is from a known class, then the identity of the instance is queried through classification.
	
	The former process is commonly known as novelty detection (or novel class detection) \cite{Markou03noveltydetection} in the literature. Given a set of image instances from known classes, the goal of novelty detection is to determine whether an observed image during inference belongs to one of the known classes. Novelty detection is generally a more challenging task than out-of-distribution detection \cite{OOD} since novel object samples are expected to be from a similar distribution to that of known samples.
	

	\begin{figure}[t]
		\centering
		\includegraphics[width=\linewidth]{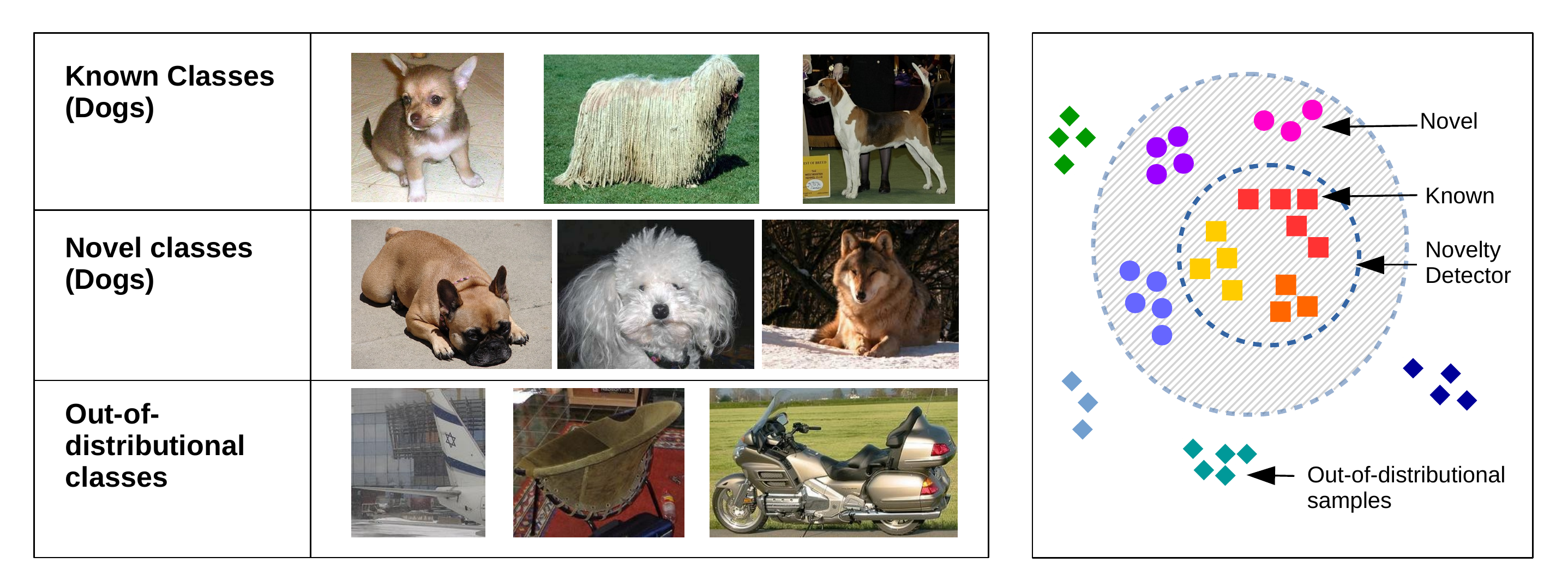}\hskip30pt
		\caption{Novelty detection in dog-breed classification. \textbf{Left:} Sample images.  \textbf{Right:} Feature representation. Both known (first row) and novel (second row) images are images of dogs. Given known images, the goal of novelty detection is to reject novel images. In order to do so, the knowledge of out-of-distributional images (final row), in this case non-dog images, are used to learn a suitable representation. }
		\label{fig:negative}
	\end{figure}

	In practice, the knowledge on unknown classes is not entirely absent. Given a set of known classes from a certain problem domain, generally unknown class data from the same problem domain is unavailable. However, in some cases it is possible to obtain data outside the known class from different problem domains, which we refer to as \textit{out-of-distributional} samples. For example, for a dog-breed recognition application, ImageNet dataset \cite{ILSVRC} that contains images of objects may be considered as \textit{out-of-distributional} data as shown as in Figure~\ref{fig:negative}.  However, since the  \textit{out-of-distributional} data are from a different  problem domain, they do not approximate the distribution of the \textit{novel} samples well.
	
	Nevertheless, since the deep-models produce generalizable features, the knowledge of \textit{out-of-distributional} samples can be transferred to the original problem to aid novelty detection. When the problem considered is a $c$ class problem, and when the \textit{out-of-distributional} data of $\mathcal{C}$ classes are available, the following three strategies are used to transfer knowledge for novelty detection in the literature:
	
	\noindent1. \textbf{Fine-tuning}: Network is first pre-trained on the \textit{out-of-distributional} data and later fine-tuned on the training data of the given domain. Novelty is queried by thresholding the final activation score \cite{BendaleB16}. 
	
	\noindent2. \textbf{Feature Extraction}:   Conventional novelty detection techniques \cite{Bodesheim_2013_CVPR},\cite{Liu},\cite{kextremes} are used based on the fine-tuned features.
	
	\noindent3. \textbf{Fine-tune $(c+\mathcal{C})$}: Network is first pre-trained on the \textit{out-of-distributional} data. Both the training data and the \textit{out-of-distributional} data are used to perform fine-tuning in $(c+\mathcal{C})$ classes together. Novelty is determined in the same way as in approach 1.

	We note that in all these baselines, the  \textit{out-of-distributional} data is employed in the training process. In fact, any novelty detection method operating on the pre-trained/finetuned deep features are implicitly making use of  the \textit{out-of-distributional} data. In this work, we introduce a new framework to perform novelty detection based on transfer learning. First, we show that using cross-entropy loss alone for training is not optimal for the novelty detection task. Secondly, we empirically show that the \textit{out-of-distributional} data can be used more effectively in training to produce better novelty detection performance with respect to considered baseline solutions. Specifically, we make following primary contributions in this paper.

	\noindent1. We propose an end-to-end novelty detection framework based on deep learning. To the best of our knowledge, this is one of the first end-to-end deep learning solutions targeting visual novelty detection.
	
	\noindent2. We introduce a new loss function, \textit{membership loss} which has a similar functionality to that of the cross-entropy loss but encourages an embedding that produces high activation for known object classes consistently.
	
	\noindent3. We propose to take advantage of large-scale external datasets to learn the \textit{globally negative filters} to reduce high activations caused by the novel images.
	
	\noindent4. We show empirically, that the proposed method outperforms the baseline novelty detection methods across four publicly available datasets.

	\section{Related Work}
	
	Object classification schemes are often equipped with a suitable mechanism to detect novel objects. For example, Eigenfaces \cite{Turk1991} was accompanied by a reconstruction error-based novel object detection method. In sparse representation-based classification (SRC) algorithm \cite{Wright:2009:RFR:1495801.1496037}, Sparsity Concentration Index (SCI) was proposed for the same purpose. 
	In contrast, there is no formal novelty detection mechanism proposed for deep-learning based classification. In its absence, thresholding the highest class activation score of the deep model has been used as a baseline in the literature \cite{BendaleB16}. As an alternative, several recent works have proposed novelty detection schemes based on deep features \cite{BendaleB16},\cite{kextremes}. In the same spirit, it is also a possibility to use classical novelty detection tools such as Kernel PCA \cite{HOFFMANN2007863}, Kernel null space-based novelty detection (KNFST) \cite{Bodesheim_2013_CVPR} and its variants \cite{localnovelty},\cite{Liu} on deep features. KNFST operating on deep-features produces the current state of the art performance in visual novelty detection \cite{Liu}. However, advantages of deep-learning are not properly exploited in all of these approaches due to the absence of an end-to-end learning framework.
	

	On the other hand, novelty detection problem has a close resemblance to both anomaly detection \cite{oza2019one}, \cite{2018arXiv180105365P}, \cite{Chandola:2009:ADS:1541880.1541882},\cite{oza2019active} and open-set recognition problems \cite{Scheirer_2013_TPAMI},\cite{BendaleB16}. Therefore, it is possible to solve anomaly detection using tools proposed in these alternative domains.  In anomaly detection, given a single \textit{normal} class, the objective is to detect out-of-class instances. One-class SVM \cite{Scholkopf:2001:ESH:1119748.1119749} and SVDD \cite{Tax:2004:SVD:960091.960109} are two of the most widely used tools in anomaly detection. Novelty detection can be viewed as an anomaly detection problem if all known classes are considered as a single augmented class. On the other hand, objective in open-set recognition (OSR)  is similar to that of novelty detection. But in addition, OSR requires correct classification of samples detected as known samples. Therefore, it is also possible to use open-set recognition tools to perform novelty detection. However, we note that due to subtle differences in objectives, OSR algorithms are not optimal for novelty detection. 
	
	In the proposed framework, maximal activation of the final layer of a deep network is considered as a statistic to perform novelty detection. We design the network and choose loss functions appropriately so that this statistic is low for novel objects compared to the known object classes. 
	\section{{Background}}
	

	\label{mechanics}
	In this section, we briefly review how deep networks produce activations in response to input stimuli. Based on this foundation, we introduce the notion of \textit{positive filters} and  \textit{negative filters}. Consider a $c$ class fully-supervised object classification problem with a training image set $\mathbf{x} = {{x_1},{x_2},\dots,{x_n}}$ and the corresponding labels $\mathbf{y} = {{y_1}, {y_2}, \dots ,{y_n}}$ where $y_i \in \{1,2, \dots c\}$. Deep convolutional neural networks (CNNs) seek to learn  a hierarchical, convolutional filter bank with filters that respond to visual stimuli of different levels. In $c$ class classification, the top most convolutional filter activation $\mathbf{g}$ is subjected to a non-linear transformation to generate the final activation vector $\mathbf{f} \in \mathbb{R}^{c}$ (for example, $\mathbf{g}$ is the conv5-3 layer in VGG16 and conv5c in Resnet50. $\mathbf{f}$ is the fc8 and fc1000 layers in the respective networks). In a supervised setting, network parameters are learned such that  $ \arg\max \mathbf{f} = {y_i}$ for $ \forall i \in \{1,2,\dots ,n \}$. This is conventionally done by optimizing the network parameters based on the cross-entropy loss.

	If there exist $k$ filters in the top most convolution filter bank, its output $\mathbf{g}$ is a set of $k$ number of activation maps. The final activation vector of the network $\textbf{f}$ is a function of $\mathbf{g}$. For a given class $i$, there exists some $k_i$ filters in the filter bank ($1 \leq k_i \leq k$) that generally generates positive activation values. These activations provide supporting (positive) evidence that an observed image is from class $i$. Conversely, all remaining filters provide evidence against this hypothesis. Activation score of each class in $\textbf{f}$ is determined by taking into account the evidence for and against each class. For the remainder of the paper, we call filters that provide evidence for and against a particular class as \textit{positive filters} and \textit{negative filters} of the class, respectively.

	\begin{figure}[t]
		\centering
		\includegraphics[width=1\linewidth]{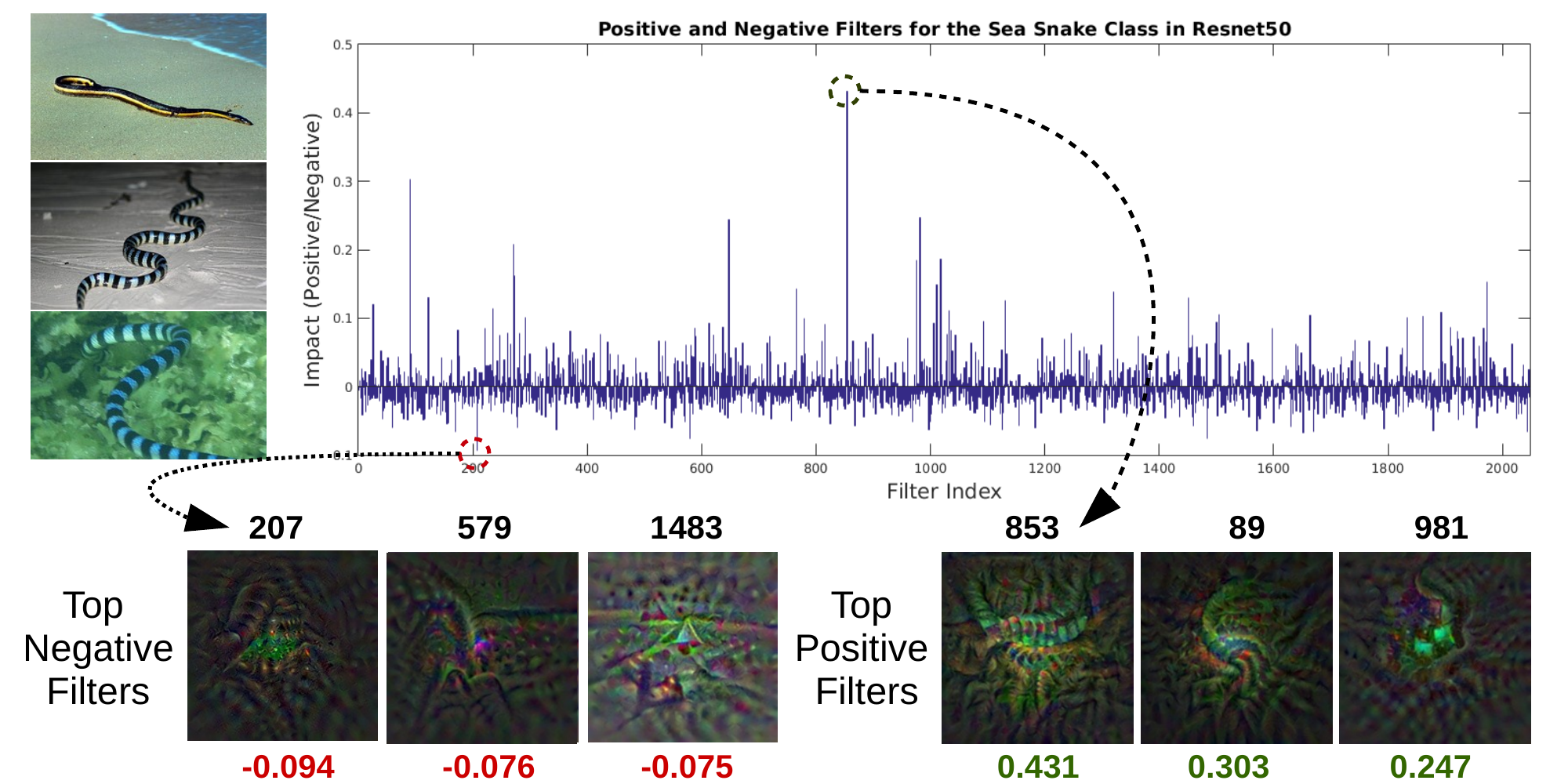}\hskip30pt
		\caption{Positive and negative filters of the \textit{sand snake} class in the Resnet50 trained on ILSVRC12 dataset. Top: weights of the fully connected layer corresponding to the  \textit{sand snake} class. We call filters associated with positive weights as positive filters of \textit{sand snake} class. All other filters are named as negative filters. Bottom: Visualization of top negative and positive filters. These patterns are likely to produce high activation in these filters. We note top positive filters are activated by snake-like structures. }
		\label{fig:snake}
	\end{figure}

	This concept can be easily explained by taking the Resnet architecture \cite{he15deepresidual} as an example. In Resnet, final convolution output $\mathbf{g}$ is subjected to global average pooling followed by a fully connected layer. Therefore, the $i^{th}$ component of the final activation vector $\mathbf{f}$ can be written as $\mathbf{f}_i =  W_i \times GAP(g)$, where $GAP$ is global average pooling operation (mean of filter map) and $W$ is the weight matrix of the fully connected layer. Here, activation of the $i^{th}$ class is a weighted summation of mean feature maps found in $\mathbf{g}$. From the above definition, filters associated with positive weights for a given class in $W$ can be identified as \textit{positive filters} for that particular class. Conversely, filters associated with the negative weights become \textit{negative filters} of the class. 
	
	For example consider the \textit{Sand Snake} class appearing in the ILSVRC12 dataset \cite{ILSVRC}. Shown in Figure~\ref{fig:snake} (top) are the weights associated with the \textit{Sand Snake} class in the final fully connected layer of the Resnet50 network trained on the ILSVRC12 dataset. We recognize filters associated with positive and negative weights as positive and negative filters, respectively for the given class. In Figure~\ref{fig:snake} (bottom) we visualize per-unit visualization of top positive and top negative filters for the considered class using the DeepVis toolbox \cite{yosinski-2015-ICML-DL-understanding-neural-networks} (these are the images that are most likely to activate the corresponding filters). By observation, we notice that the top \textit{positive filters} are activated when the network observes structures similar to snakes. On the other hand, the top \textit{negative filters} are unrelated to the appearance of \textit{sand snakes}.

	\section{Deep Novelty Detection}
\label{obj}
	Based on the above background, we propose to learn the distributions of known object classes using a CNN framework with the objective of performing joint classification and novelty detection. In our formulation, assuming each known class has a unique single label, we force the final activation vector  $\mathbf{f}$ to model the probability distribution vector of known classes. Formally, for a given data-label pair $(x_i,y_i)$, we expect $\mathbf{f}_i=1$ and $\mathbf{f}_j=0,~ \forall j\neq i$. Once such a representation is learned, $\arg \max \mathbf{f}$ returns the most-likely class of an observed sample. Then, $\max \mathbf{f}$ yields the likeliness of the sample belonging to the most likely class. Similar to binary classification,  identity $I$ of a test instance can be queried using hard thresholding. In order to learn a representation suitable for the stated objective, we use conventional classification networks as the foundation of our work and propose the following two alternations.

	
	\noindent \textbf{1. Membership loss.} Assuming each known class has a unique single label, if the observed image is from a known class, only a single positive activation should appear in $\mathbf{f}$. We observe that when cross-entropy loss is used, this is not the case. To alleviate this, we introduce a new loss called \textit{{membership loss}} in addition to the cross-entropy loss. 
	
	\noindent \textbf{2. Globally negative filters.} In a classification setting, a negative filter of a certain class is also a positive filter of another class. In other words, there exist no explicit negative filters. In our formulation, we propose to generate \textit{globally negative filters} (filters that generate negative evidence for all known classes) to reduce the possibility of a novel sample registering high activation scores.

	

	\begin{figure*}[tbh!]
		\centering
		\includegraphics[width=1\linewidth]{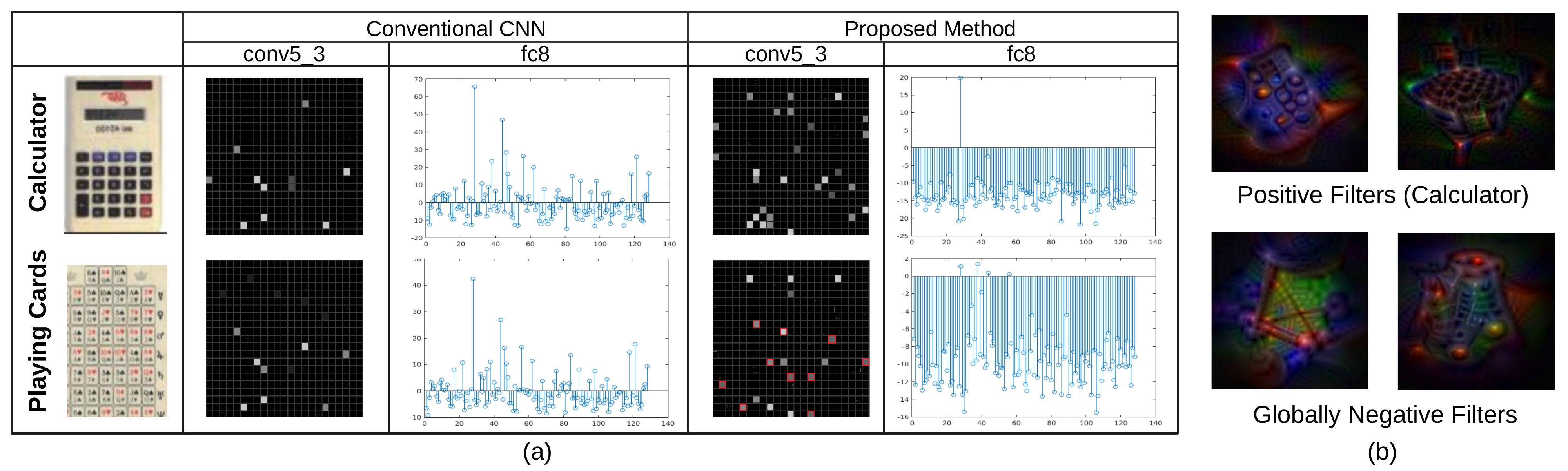}\hskip30pt
		\caption{(a) Activations of known (Calculator) and unknown samples (Playing Cards) in a VGG16 model. In conventional CNN, both known and unknown samples activates similar conv5-3 filters and results in a similar fc8 activation map. Novelty detection of the novel sample fails due to high activation scores present in fc8 layer. In the proposed method, Calculator object activates filters related to  Calculators whereas top activated filters in Playing Cards is unrelated to known classes (globally negative). Since all activations in fc8 are very small for the Playing Cards object, it can be detected as a novel sample by thresholding. (b) Top positive filters and top globally negative filters of the calculator class.}
		\label{fig:acti2}
	\end{figure*}
	
	\subsection{{Limitations of Cross-Entropy Loss}}
	When a classification network is trained, each element $f_i$ of the activation vector $\mathbf{f}$ is first normalized using the softmax function to arrive at a normalized activation vector $\mathbf{\tilde{f}}$ as in,
	$\tilde{f_j} =  e^{{f}_j} / {\sum\limits_{j=1}^{c}e^{{f}_j}}.$  When it is assumed that all image classes appearing during inference are known ahead of time, $j^{th}$ element of vector $\tilde{\mathbf{f}}$ is interpreted as the likelihood of the input image $x_i$ belonging to the $j^{th}$ class. Neural network-based classification systems are learned by minimizing the cross-entropy loss which is the negative log likelihood of the correct class $\tilde{\mathbf{f}}$. However, since this is a relative measure, the learned representation deviates from our objective due to the following reasons.
	
	Firstly, even a low activation of the ground truth class could yield a low cross-entropy provided that the activations of all other (non-matching) classes are very low. As a result, lower score values may not get heavily penalized during training. Therefore, a model trained using the cross-entropy loss may end up producing low activation scores for known classes during inference. In closed set classification, this behavior will not cause complications as long as the correct class records the  highest score. However, in threshold-based novelty detection, this poses a problem as having low scores for the positive class will result in false negatives. Secondly, the cross-entropy loss does not penalize activations of unrelated classes as long as the correct class produces the highest activation. As a result, inaccurate cross-class relationships are encouraged during training.

	In order to illustrate this point, we trained a VGG16 \cite{DBLP:journals/corr/SimonyanZ14a} based CNN classification network using the first 128 classes of the Caltech256 dataset. For the considered example, the Calculator class (indexed at 27) is a known  class and the Playing Cards class (indexed at 163) is a novel class. Shown in Figure~\ref{fig:acti2} are the activations of conv5-3 and fc8 layers of the network for two inputs of the two classes. As can be seen from this figure, when the network observes a calculator object (known object), it correctly associates the highest score in \textbf{f} to the correct class (class 27). However, there is also a significant miss-association between the calculator class and coin (class 43), keyboard (class 45), dice (class 55) and joystick classes (class 120).

	\subsection{Membership Loss}
	In our approach, we first independently translate each activation score value $f_i$ into the range $0 - 1$  using the sigmoid$(\sigma)$ function. We interpret each transformed activation score as the probability of the input image belonging to each individual class. If the ground truth label of a given observation $x$ is $y$, we aim at learning a function that produces absolute probabilities for the membership of each class as follows
	\begin{equation}\mathbb{P}(y=i) = \sigma(f(x)_i) ~~\forall i \in \{1,2,\dots c\}.\end{equation}
	Ideally, the learned transformation will produce $f(x)_i = 1 $ for $i=y$ and $f(x)_i = 0,$ otherwise. We denote the risk of associating a higher score with a wrong class ($f(x)_i = 1 $ for $i \neq y$ ) as  $R_{W1}$ and risk of associating a low score with the correct class ($f(x)_i = 0 $ for $i=y$) as  $R_{C0}$. We define the \textit{membership loss $L_M$} as the risk of classification as  
	\begin{equation}
	\nonumber L_M(x,y) = R_{C0}(x,y) + \lambda R_{W1}(x,y),\end{equation}
	where $\lambda$ is a positive scalar. With our formulation, we define   $R_{W1}(x,y) = [1-\mathbb{P}(y=1)]^2 = [1-\sigma(f(x)_y)]^2$. Here, the quadratic term is introduced to impose a heavy penalty on very high deviations. Similarly, $R_{C0}(x,y)$ becomes,
	\begin{equation}
	\begin{aligned}
	\nonumber R_{C0}(x,y) &= \frac{1}{c-1}\sum\limits_{i=1, i\neq y}^{c}[\mathbb{P}(i=1)]^2\\ &= \frac{1}{c-1}\sum\limits_{i=1, i\neq y}^{c} [\sigma(f(x)_i)]^2.\end{aligned}\end{equation}
	By substitution, we get	
	\begin{equation}\nonumber	L_M(x,y) = [1-\sigma(f(x)_y) ]^2+ \lambda  \frac{1}{c-1}\sum\limits_{i=1, i\neq y}^{c} [\sigma(f(x)_i)]^2. \end{equation}
	Here, the parameter $\lambda$ controls relative weight given to each risk source. In our experiments, we set $\lambda = 5$. Taking the partial derivative of the membership loss yields the following back-propagation formula
	\begin{equation}	
\nonumber	\frac{\partial L_M(x,y) }{\partial f(x)_i}=
	\begin{cases}
	-2[1-\sigma(f(x)_i)] \times \sigma(f(x)_i)'& \text{for } i=y\\
\nonumber	\frac{2\lambda}{c-1} \sigma(f(x)_i) \times \sigma(f(x)_i)'& \text{for } i\neq y,
	\end{cases}
	\end{equation}
	where, $\sigma(f(x)_i)' = \sigma(f(x)_i) (1-\sigma(f(x)_i))$. 
	
	The proposed \textit{membership loss} does not operate on the closed-set assumption. It takes individual score values into account in an absolute sense. Therefore, when the membership loss is used, known samples that produce small activations will be penalized regardless of the score values of the other classes. When the membership loss is used together with the cross-entropy loss, the network learns a representation that produces relatively higher activation scores for the correct class. For example, consider the $fc8$ activation map of the proposed method for the Calculator object input shown in Figure~\ref{fig:acti2}. There, we observe that the correct class (indexed at 27) produces a large positive score whereas all other classes produce negative scores.

	\subsection{Globally Negative Filters}
	
	When a conventional classification network is used, novel images are often able to produce very high activation scores there by leading to false positive detections. Such an example is shown in Figure~\ref{fig:acti2}(bottom) where a Playing Cards instance has produced a very high activation score in the index corresponding to the Calculator class (indexed at 27). Final activation score of a class is generated based on the responses of the positive and negative filters as discussed in Section~\ref{mechanics}. Once the network is trained, given an input of a particular known class, the input stimulates some \textit{positive filters}  and \textit{negative filters} associated with the class. If the model is well trained, the response of the positive filters exceeds the response of the negative filters to produce a high positive activation score. 
	
	Given this background, it is interesting to investigate how a novel sample is able to produce a high activation score. Let us revisit activations of Playing Cards image (novel image) shown in Figure~\ref{fig:acti2} (bottom). In this example, Playing Cards image has stimulated some positive filters of the Calculator class despite the differences in content. At the same time, by chance, it has not produced sufficient stimulation in negative filters of the Calculator class, thereby producing a large positive activation in $\mathbf{f}$. This can be clearly observed in Figure~\ref{fig:acti2} where both the Calculator and the Playing Cards images have activated similar filters in the conv5-3 layer. 
	
	
	To this end, we make the following proposal. We wish to learn a set of filters that are stimulated generally by natural images and produce evidence against all known classes. In other words, these filters are \textit{negative filters} with respect to all known classes - hence we call them \textit{globally negative filters}. If any of such filters are stimulated during inference, it would prove greater evidence that the observed image is novel. However, this proposal will succeed only if the \textit{globally negative filters} are stimulated by arbitrary images outside the known class set. 
	
	In order to learn the \textit{globally negative filters}, we propose a joint-learning network structure. In addition to the known object dataset, we use the \textit{out-of-distributional} data samples in training. For the remainder of the paper we refer the \textit{out-of-distributional} dataset as the \textit{reference dataset}. We learn features that can perform classification in both the known dataset and the reference dataset. If the \textit{reference dataset} has $\mathcal{C}$ classes, once trained, the filter bank will contain positive filters of all $c+\mathcal{C}$ classes. Filters associated with the reference dataset will likely act as \textit{negative filters} for all classes in the known dataset, thereby be globally negative. In this framework, the \textit{globally negative filters} are likely to respond to arbitrary natural images provided that the reference dataset is a  large-scale diverse dataset. 
	
	In Figure~\ref{fig:acti2}, we show the impact of using the \textit{globally negative filters}. Visualization of top activated filters for the Calculator class are shown at the top in Figure~\ref{fig:acti2}(b). As can be seen from this figure, these filters are positively co-related with the Calculator class. With the new formulation, we observe that playing cards object activates some extra filters which are not in common with the calculator class (highlighted in red). At the bottom of Figure~\ref{fig:acti2}(b) we visualize filters with the highest activation for the Playing Cards object. By inspection, these two visualizations look arbitrary and do not have an obvious association with any of the Caltech256 classes. We interpret these filters as instances of the \textit{globally negative filters}. Due to the availability of more negative evidence, the overall activation value of the playing cards object has been drastically reduced. 
	
	\subsection{Training Procedure}
	We propose a network architecture and a training mechanism to ensure that the network learns the \textit{globally negative filters}. For this purpose, we use an external multi-class labeled dataset which we refer to as the \textit{reference dataset}. 
	
\begin{figure}[tbh!]
		\centering
		\includegraphics[width=1\linewidth]{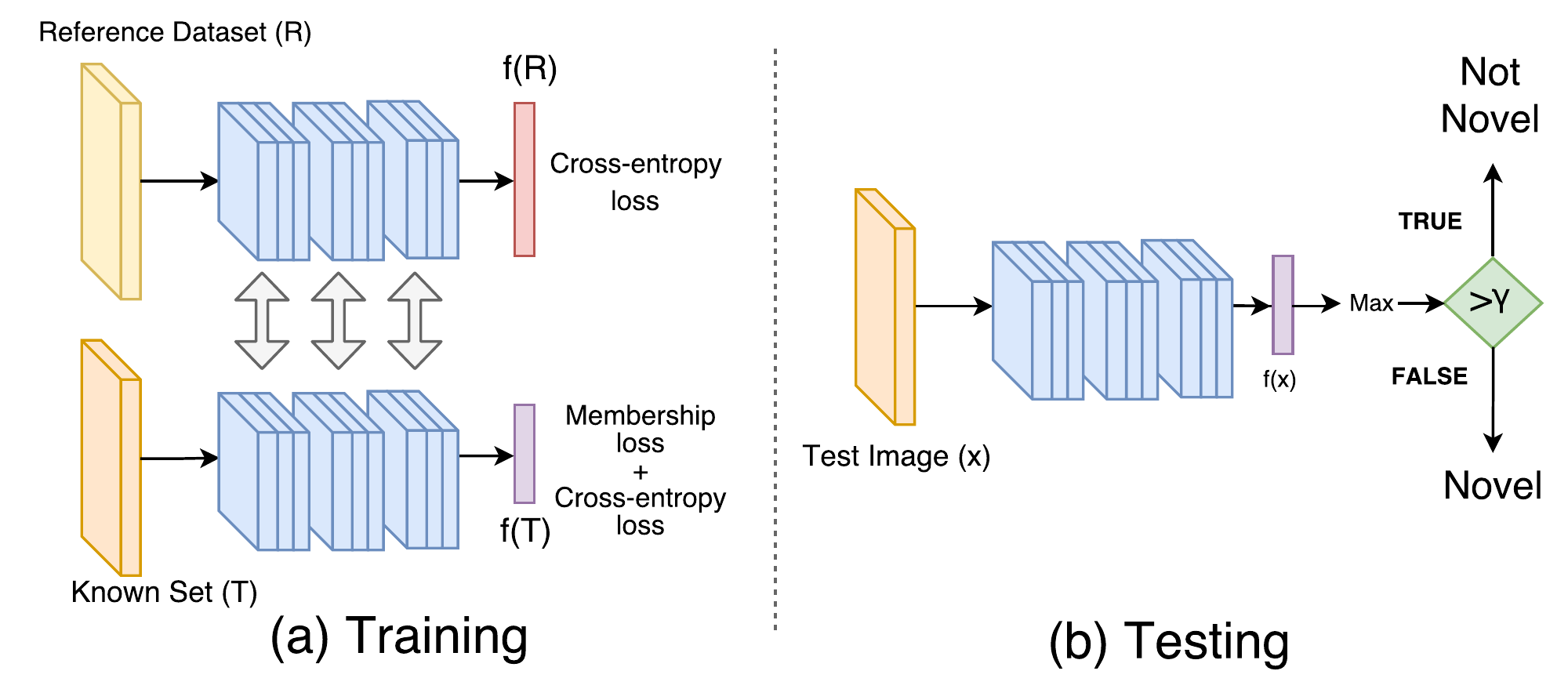}\hskip30pt
		\caption{Proposed architecture for novelty detection. We use an external multi-class dataset (reference dataset (R)) in addition to the known object dataset (T). Two parallel CNN networks with identical structure and weights are used to extract features from both datasets. We train separate classifier networks operating on the same feature to perform classification in either dataset. During inference, novelty detection is performed by thresholding the maximal activation of the bottom branch of the network.}
		\label{fig:nw}
	\end{figure}

	We first select a CNN backbone of choice (this could be a simple network such as Alexnet \cite{NIPS2012_ALEX} or a very deep/complex structure such as DenseNet \cite{huang2017densely}). Two parallel CNN networks of the selected backbone are used for training as shown in Figure~\ref{fig:nw}(a). The only difference between the two parallel networks is the final fully-connected layer where the number of outputs is equal to the number of classes present in either dataset. For the purpose of our discussion, we refer the sub-network up to the penultimate layer of the CNN as the feature extraction sub-network.

	Initially, weights of the two feature extraction sub-networks are initialized with identical weights and they are kept identical during training. Final layer of both parallel networks are initialized independently. Weights of these two layers are learned during training without having any dependency between each other. During training, two mini batches from two datasets (reference dataset (R) and known classes (T)) are considered and they are fed into the two branches independently. We calculate the cross-entropy loss $(L_{ce})$ with respect to the samples of the reference dataset and both the membership loss $(L_{m})$ and the cross-entropy loss with respect to the samples of known classes. The cumulative loss of the network then becomes a linear combination of the two losses as follows,
	\begin{equation}\nonumber 
	Cumulative Loss = L_{ce}(R)+\alpha_1~L_{ce}(T)+\alpha_2~L_{m}(T).\end{equation}
	
	In our experiments, we keep $\alpha_1, \alpha_2 = 1$. The cumulative loss is back-propagated to learn the weights of the two CNN branches. Reducing membership loss and cross-entropy loss with respect to the known-class dataset increases the potential of performing novelty detection in addition to classification as discussed in the preceding sub-section. On the other hand, having good performance (low cross-entropy loss) in the reference dataset suggests the existence of filters that are responsive to generic objects provided that the reference dataset is sufficiently diverse. When classes appearing in the reference dataset do not intersect with known classes, these filters serve as the \textit{globally negative filters}.

	\subsection{Testing (Novelty Detection)}
	During inference, we propose to use the setup shown in Figure~\ref{fig:nw}(b) where we only consider the bottom CNN branch of the training network. Given a test image $x$, we perform a forward pass using the learned CNN network to obtain the final feature $\textbf{f}(x)$. The largest element of $\textbf{f}(x)$, $\max \textbf{f}(x)$ is thresholded using a predetermined threshold $\gamma$ to arrive at the identity of the test image. If the yielded score is below the threshold $\gamma$, we identify the test sample to be novel. In a practical system, threshold $\gamma$ is chosen considering the percentile of the matched score distribution (for example threshold can be chosen to be 95th percentile if the accepted false negative rate is 5\%) . In addition to the novelty detection procedure, the same network structure can be used to perform classification as well. Here, $\arg \max \textbf{f}(x)$ yields the identity of the predicted class for the test sample $x$. We note that this step is identical to the classification procedure used in the standard CNN-based classification.

	\section{Experimental Setup and Results}
	In this section, we present experimental results for the novelty detection task. We first describe the baseline methods used for comparison. Then, we introduce the four datasets used for evaluation. Finally, we discuss the obtained results followed by the analysis of the proposed method.
	
	\subsection{Baseline Methods} 
	We evaluate the proposed method on four novelty detection databases and we compare its performance with the standard novelty detection schemes. We use the  following baseline comparisons based on the AlexNet \cite{NIPS2012_ALEX} and the VGG16 \cite{DBLP:journals/corr/SimonyanZ14a} features fine-tuned on the given dataset. 
	
		\noindent\textbf{1. Finetune \cite{DBLP:journals/corr/SimonyanZ14a}:}  $fc8$ feature scores of the trained deep model are thresholded to detect novel samples.\\
	\noindent\textbf{2. One-class SVM \cite{Scholkopf:2001:ESH:1119748.1119749}:} A one-class SVM classifier is trained for all known classes. The maximum SVM score is considered during the inference.\\
		\noindent \textbf{3. KNFST \cite{Bodesheim_2013_CVPR}, \cite{Liu}:} Deep features are normalized and histogram intersection kernel method is used to generate inner products between the samples.\\
		\noindent \textbf{4. Local KNFST \cite{localnovelty}:} Deep features with histogram intersection kernel is considered with 600 local regions.\\
		\noindent \textbf{5. OpenMax \cite{BendaleB16}:} Activations of penultimate layer of a deep model are used to construct a single channel class-wise mean activation vectors (MAV) and the corresponding Weibull distributions.\\
		\noindent \textbf{6. K-extremes \cite{kextremes}:} Mean activations of the VGG16 $fc7$ features are considered for each class and top 0.1 activation indexes are binarized to arrive at the Extreme Value Signatures.\\
		\noindent \textbf{7. Finetune$(c+\mathcal{C})$:} A $(c+\mathcal{C})$ class CNN is trained by treating classes of the reference dataset as the additional class.
		\\
		In addition, we evaluate the performance based on the pretrained deep features (trained on the ILSVRC12 database) for KNFST and local KNFST methods. Whenever pre-trained features are use they are denoted by the suffix \textit{pre}.

	\subsection{Datasets} 
	We use four publicly available multi-class datasets to evaluate the novelty detection performance of the proposed method.

	\begin{figure}[tbh!]
		\centering
		\includegraphics[width=1\linewidth]{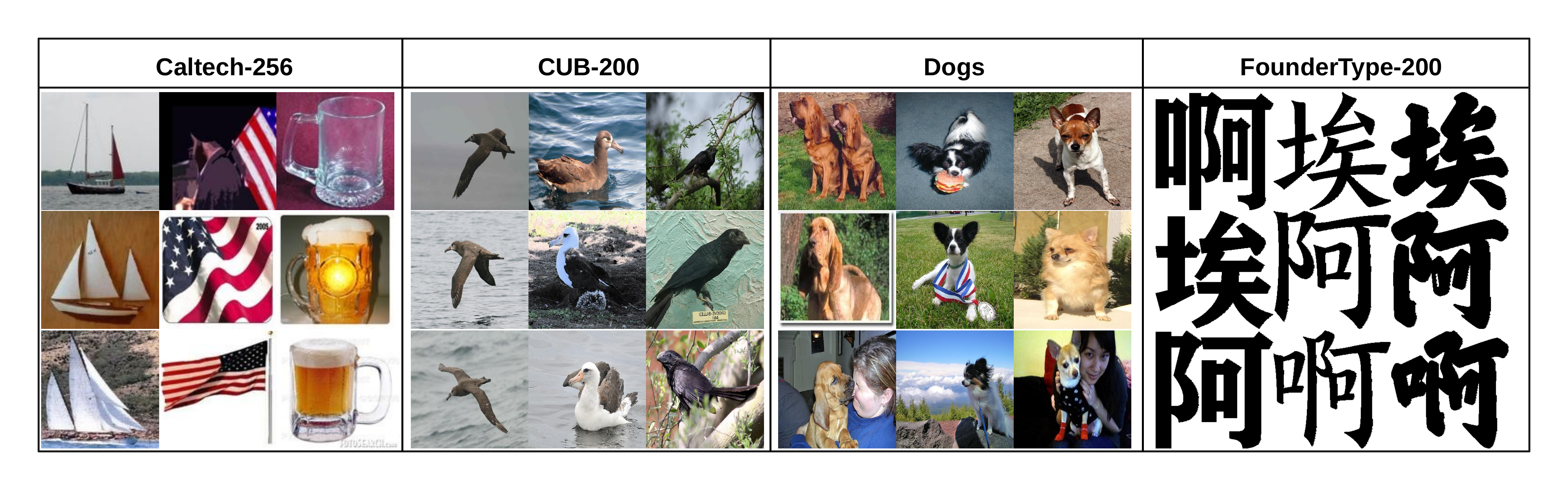}\hskip30pt
	\vskip -5pt	\caption{Sample images from the evaluation datasets. Each column contains images taken from a single class of each dataset.}
		\label{fig:samples}
	\end{figure}

	\noindent\textbf{Caltech256 Dataset.} The Caltech256 dataset is a fully annotated dataset which consists of 30607 images from 256 object classes. Following the protocol presented in \cite{Liu}, we first sorted the class names alphabetically and picked the first 128 classes as the known classes and considered the images from the remaining 128 classes as the novel images. 
	
	\noindent\textbf{Caltech-UCSD Birds 200 (CUB 200) Dataset.} 
	The CUB-200 dataset includes 6033 images belonging to 200 distinct bird categories. Ground truth labels for each image are provided. In our experiment, we sorted names of the bird categories alphabetically and used the first 100 classes as the known classes.  The remaining classes were used to represent novel images.

	\noindent\textbf{Stanford Dogs Dataset.} 
	This dataset is a subset of the ImageNet dataset and was originally intended for fine-grain classification. There are 20580 images belonging to 120 different dog breeds in this dataset. We considered the first 60 classes as the known classes and treated the remaining classes as the novel classes during performance evaluation.

	\noindent\textbf{FounderType-200 Dataset.} 
	This dataset is a collection of Chinese character images in different font types. The dataset is organized based on the font-type. In total there are 200 different font-types with 6763 images from each class in this dataset. Following the same convention as before, we picked the first 100 classes to represent the enrolled classes. The remaining 100 classes were used to simulate the novel images. 
	
	In all datasets, following the protocol in \cite{Liu}, images of the enrolled classes were randomly split into two even sets to form training and testing datasets of the enrolled classes. Images of the novel classes were used only during testing. When finetuning/extracting features from the caltech256 dataset  following \cite{imagenet_cvpr09}, we used the pretrained model trained on the Places365 dataset \cite{zhou2017places}. For all other tasks, we used the pretrained model trained on the ILSVRC12 dataset. Accordingly, the validation sets of Places365 was used as the reference dataset for Caltech256. For all other tasks the validation set of ILSVRC12 was considered. 
	
	\subsection{Results} \label{sec:results}
	We evaluated all methods based on the VGG16 and the AlexNet features.  We used the training codes provided by the authors when evaluating the KNFST \cite{Bodesheim_2013_CVPR} and the local KNFST \cite{localnovelty} methods. Performance of each method is evaluated using the area under the receiver operating characteristics (AUC) curve. Obtained AUC values for each method are tabulated in Table~\ref{table:nov} for all datasets\footnote{Source code of the proposed method is made available at https://github.com/PramuPerera/TransferLearningNovelty}.

	\begin{table}[htp!]
		\centering
		\caption{Novelty detection results (AUC of the ROC curve) on the evaluation datasets. The best performing method for each dataset is shown in bold. Second best method is shown in italics.}
		\label{table:nov}
		\resizebox{1\linewidth}{!}{
			
			\begin{tabular}
				{|l|p{1.2cm}|p{1.2cm}|p{1.2cm}|p{1.2cm}|p{1.2cm}|p{1.2cm}|p{1.2cm}|p{1.2cm}|}
				\hline
				
				Method&				 \multicolumn{2}{|c|}{Caltech-256} & 				 \multicolumn{2}{|c|}{CUB-200} &  				 \multicolumn{2}{|c|}{ 
					Dogs} & 				 \multicolumn{2}{|c|}{FounderType}\\ \hline
				& VGG16 & AlexNet & VGG16 & AlexNet & VGG16 & AlexNet & VGG16 & AlexNet  \\ \hline
				Finetune\cite{DBLP:journals/corr/SimonyanZ14a}, \cite{NIPS2012_ALEX} & 0.827 & 0.785 & 0.931 & 0.909 & 0.766 & 0.702 & 0.841 & 0.650
				\\
				One-class SVM\cite{Scholkopf:2001:ESH:1119748.1119749} & 0.576 & 0.561 & 0.554 & 0.532& 0.542 & 0.520 & 0.627 & 0.612\\
				KNFST pre\cite{Bodesheim_2013_CVPR}& 0.727 & 0.672 & 0.842 & 0.710 & 0.649 & 0.619 & 0.590 & 0.655
				\\
				KNFST\cite{Bodesheim_2013_CVPR}, \cite{Liu}& 0.743 &  0.688  & 0.891 & 0.748 & 0.633 & 0.602 & \textit{0.870} &0.678
				\\
				Local KNFST pre\cite{localnovelty}& 0.657 & 0.600 
				&  0.780 &  0.717 & 0.652 & 0.589 & 0.549 &  0.523\\
				Local KNFST\cite{localnovelty}& 0.712 &  0.628&  0.820 & 0.690 & 0.626& 0.600& 0.673 & 0.633\\
				K-extremes\cite{kextremes}& 0.546 & 0.521
				& 0.520 & 0.514& 0.610 & 0.592& 0.557 & 0.512\\
							OpenMax\cite{BendaleB16} & 0.831 & 0.787 & \textit{0.935} & \textit{ 0.915} & 0.776 & \textit{0.711} & 0.852 & 0.667\\
				Finetune$(c+\mathcal{C})$& \textit{0.848} & \textit{0.788}
				& 0.921  & 0.899 & \textit{0.780} & 0.692& 0.754  & \textit{0.723}\\

				
				Deep Novelty (ours) & \textbf{0.869 } &  \textbf{0.807} & \textbf{0.958} & \textbf{0.947} & \textbf{0.825} & \textbf{0.748} & \textbf{0.893} & \textbf{0.741} \\
				\hline
			
			\end{tabular} 
		}

	\end{table}
	
	When baseline methods are considered, a variance in performance can be observed across datasets. In general, K-extremes has reported below-par performances compared to the other methods. When the number of enrolled classes are very high, the mean activation signature of a class looses its uniqueness. This is why K-extremes method fails when very large number of classes are enrolled as suggested in \cite{kextremes}. In the Caltech-256 and CUB-200 datasets, thresholding deep activations and OpenMax has yielded better results among the baseline methods. In Caltech256, this has improved marginally when the reference dataset (ILSVRC12) is incorporated. This method has performed reasonably well in the FounderType-200 dataset but it's performance in the Standford Dogs dataset is not convincing. In general, KNFST has out-performed local KNFST except for in the Standford Dogs dataset. KNFST (and local KNFST) operating on the finetuned deep features have performed better in general compared to the pre-trained deep features. This trend has changed only in the Standford Dogs dataset. Here we note that none of the baseline methods have yielded consistent performance across datasets.
	
	
	In comparison, the proposed method is able to produce the best performance across all datasets. When AlexNet is used as the back-bone network, there is an improvement of about 3.0\% over the baselines in the CUB-200 and Standford Dogs datasets. In the other two datasets this margin is 2.0\%. In the Caltech256, CUB-200 and FounderType-200 datasets, the improvements in AUC are in excess of 2.0\% for the VGG16 model. In the Standford Dogs dataset, the proposed method is able to introduce a significant advancement of more than 7.0\% in AUC compared with the baseline methods. In general, we note that in datasets where the baseline performance is already very good, as in the CUB-200 and FounderType 200 datasets, the improvement of the proposed method is relatively small. On the other hand, when the baseline performance is poor, the proposed method is able to generate a significant improvement in the performance.

	\subsection{Ablation Study}
	In this subsection, we investigate the impact of each individual component of the proposed framework. For the purpose of the ablation study, we use the validation dataset of the ILSVRC12 dataset as the reference dataset. It should be noted that figures reported in this subsection are different from Table~\ref{table:nov} due to this reason. Starting from the traditional CNN architecture, we added one component of the proposed framework at a time and evaluated the novelty detection performance on the Caltech-256 dataset as a case study. Testing protocol presented in the preceding subsection was followed in all cases. Considered cases are as follows.
	
	\noindent \textbf{a) Single CNN with the cross-entropy loss (AUC 0.854).} This is the CNN baseline where a CNN is trained using the enrolled classes conventionally. 
	
	\noindent \textbf{b) Single CNN with the cross-entropy loss+membership loss (AUC 0.865).} The network architecture is the same as in case (a). In addition to the cross-entropy loss, the membership loss is calculated with respect to the enrolled dataset.
	
	\noindent \textbf{c) Two Parallel CNNs with cross-entropy loss (AUC 0.864).} The network structure proposed in Figure~\ref{fig:nw}(a) is used. In contrast, only the cross-entropy loss is used in the bottom sub-network.
	
	\noindent \textbf{d) Proposed method (AUC 0.906). } Proposed structure Figure~~\ref{fig:nw}(a) is used for training.
	
	In the proposed method, we introduced membership loss and a parallel network structure as contributions. From the case study conducted, it appears that the novelty detection performance improves compared to the baseline even when one of the contributions are used. Moreover, we observe that the two contributions compliment each other and generate even better results when combined together.
	
	\subsection{Impact of the Reference Dataset}  
	In the proposed method, we assumed the availability of a reference dataset with large number of classes. In this subsection, we investigate the impact of the reference dataset by varying the reference dataset of choice. In particular, we use the ILSVRC12, Caltech-256 and Standford Dogs datasets as the reference datasets to perform novelty detection using the proposed method in the CUB-200 dataset. Results obtained are tabulated in Table~\ref{table:ref}. Here we have included the performance of the best baseline method for the CUB-200 dataset (Finetune) from Table~\ref{table:nov} as a baseline.
	
	Compared to ILSVRC12, when Caltech-256 is used as the reference dataset, AUC drops by 0.005\%. This further drops by 0.008\% when the Standford Dogs dataset is used. The ILSVRC12 dataset contains 1000 image classes and has significant variance in images within each class. Caltech-256 is a similar multi-class dataset but with fewer classes. Both of these datasets contain natural images. However since ILSVRC12 has more classes and more intra-class variance, we expect it to generate \textit{globally negative filters} better. Therefore, the performance drop of Caltech-256 compared to ILSVRC12 is expected. On the other hand, the Standford Dogs dataset only contains images of dogs. Therefore, filters learned using this dataset may not be generic to get stimulated by the arbitrary inputs. Therefore, the drop in the performance is justified. In conclusion, we note that the proposed method is able to out-perform baseline novelty detection methods even when the reference dataset is varied. However, better results are obtained when a larger dataset with high degree of intra-class variation is used as the reference dataset.

	\begin{table}[htp!]
		\centering
		\caption{Impact of the reference dataset used. Results of the case study  conducted on the CUB-200 dataset by varying the reference dataset.}
		\label{table:ref}
		\resizebox{1\linewidth}{!}{
			
			\begin{tabular}
				{|l|p{2cm}|p{2cm}|p{2cm}|p{2cm}|}
				\hline
				&Baseline &ILSVRC12& Caltech-256 &  Dogs  \\ \hline
				Novelty Detection AUC& 0.931 & \textbf{0.958} &   0.953 & 0.945
				\\
				\hline
				
			\end{tabular} 
		}

	\end{table}
	
	\subsection{Impact on Classification Accuracy}
	When a test image is present, the proposed method produces a set of class activation scores. It is still possible to perform classification using the same system by associating the test image with the class containing the highest activation. In what follows, we consider test samples of the known classes and perform closed-set classification in the same experimental setup described in Section~\ref{sec:results}. In other words, we do not consider novel samples for the purpose of this study. Obtained classification accuracies for the four datasets are tabulated in Table~\ref{table:class}. Although the proposed method is designed for the purpose of novelty detection, we note that the proposed changes have contributed towards increasing the classification accuracy of the system as well. This is because the \textit{membership loss} explicitly enforces correct class to have a high score and all other classes to have scores closer to zero.

	\begin{table}[htp!]
		\centering
		\caption{Classification accuracy obtained for conventional fine-tuning and the proposed method for the four evaluation datasets.}
		\label{table:class}
		\resizebox{1\linewidth}{!}{
			
			\begin{tabular}
				{|l|p{2cm}|p{2cm}|p{2cm}|p{2cm}|}
				\hline
				&Caltech-256 &CUB-200& Dogs &  FounderType  \\ \hline
				VGG16& 0.908 &  0.988 &  0.730  & 0.945
				\\
				\hline
				Proposed Method& \textbf{0.939} & \textbf{0.990} &  \textbf{0.801}   & \textbf{0.950}
				\\
				\hline
				
			\end{tabular} 
		}
		
	\end{table}
	
	\section{Conclusion}
	We presented an end-to-end deep learning-based solution for image novelty detection. We build up on the conventional classification networks and introduce two novel contributions; namely, \textit{membership loss} and a training procedure that produces \textit{globally negative filters}. In the proposed method, novelty is quarried simply by thresholding the highest activation of the output vector. We demonstrate the effectiveness of the proposed method on four publicly available multi-class image datasets and obtain  state-of-the-art results.
	
	
	

	
	{\small
		\bibliographystyle{ieee}
		\bibliography{egbib}

\begin{thebibliography}{10}\itemsep=-1pt

\bibitem{BendaleB16}
A.~Bendale and T.~E. Boult.
\newblock Towards open set deep networks.
\newblock In {\em 2016 {IEEE} Conference on Computer Vision and Pattern
  Recognition, {CVPR} 2016, Las Vegas, NV, USA, June 27-30, 2016}, pages
  1563--1572, 2016.

\bibitem{localnovelty}
P.~Bodesheim, A.~Freytag, E.~Rodner, and J.~Denzler.
\newblock Local novelty detection in multi-class recognition problems.
\newblock In {\em 2015 IEEE Winter Conference on Applications of Computer
  Vision}, pages 813--820, 2015.

\bibitem{Bodesheim_2013_CVPR}
P.~Bodesheim, A.~Freytag, E.~Rodner, M.~Kemmler, and J.~Denzler.
\newblock Kernel null space methods for novelty detection.
\newblock In {\em IEEE Conference on Computer Vision and Pattern Recognition
  (CVPR)}, June 2013.

\bibitem{Chandola:2009:ADS:1541880.1541882}
V.~Chandola, A.~Banerjee, and V.~Kumar.
\newblock Anomaly detection: A survey.
\newblock {\em ACM Comput. Surv.}, 41(3):15:1--15:58, 2009.

\bibitem{imagenet_cvpr09}
J.~Deng, W.~Dong, R.~Socher, L.-J. Li, K.~Li, and L.~Fei-Fei.
\newblock {ImageNet: A Large-Scale Hierarchical Image Database}.
\newblock In {\em CVPR09}, 2009.

\bibitem{he15deepresidual}
K.~He, X.~Zhang, S.~Ren, and J.~Sun.
\newblock Deep residual learning for image recognition.
\newblock In {\em IEEE Conference on Computer Vision and Pattern Recognition},
  pages 770--778, June 2016.

\bibitem{HOFFMANN2007863}
H.~Hoffmann.
\newblock Kernel pca for novelty detection.
\newblock {\em Pattern Recognition}, 40(3):863 -- 874, 2007.

\bibitem{huang2017densely}
G.~Huang, Z.~Liu, L.~van~der Maaten, and K.~Q. Weinberger.
\newblock Densely connected convolutional networks.
\newblock In {\em Proceedings of the IEEE Conference on Computer Vision and
  Pattern Recognition}, 2017.

\bibitem{NIPS2012_ALEX}
A.~Krizhevsky, I.~Sutskever, and G.~E. Hinton.
\newblock Imagenet classification with deep convolutional neural networks.
\newblock In {\em Advances in Neural Information Processing Systems 25}, pages
  1097--1105, 2012.

\bibitem{Liu}
J.~Liu, Z.~Lian, Y.~Wang, and J.~Xiao.
\newblock Incremental kernel null space discriminant analysis for novelty
  detection.
\newblock In {\em 2017 IEEE Conference on Computer Vision and Pattern
  Recognition (CVPR)}, pages 4123--4131, July 2017.

\bibitem{Markou03noveltydetection}
M.~Markou and S.~Singh.
\newblock Novelty detection: a review -- part 1: statistical approaches.
\newblock {\em Signal Processing}, 83(12):2481 -- 2497, 2003.

\bibitem{oza2019active}
P.~Oza and V.~M. Patel.
\newblock Active authentication using an autoencoder regularized cnn-based
  one-class classifier.
\newblock In {\em 2019 14th IEEE International Conference on Automatic Face \&
  Gesture Recognition (FG 2019)}. IEEE, 2019.

\bibitem{oza2019one}
P.~Oza and V.~M. Patel.
\newblock One-class convolutional neural network.
\newblock {\em IEEE Signal Processing Letters}, 26(2):277--281, 2019.

\bibitem{2018arXiv180105365P}
P.~{Perera} and V.~M. {Patel}.
\newblock {Learning Deep Features for One-Class Classification}.
\newblock {\em ArXiv e-prints}.

\bibitem{ILSVRC}
O.~Russakovsky, J.~Deng, H.~Su, J.~Krause, S.~Satheesh, S.~Ma, Z.~Huang,
  A.~Karpathy, A.~Khosla, M.~Bernstein, A.~C. Berg, and L.~Fei-Fei.
\newblock Imagenet large scale visual recognition challenge.
\newblock {\em Int. J. Comput. Vision}, 115(3):211--252, Dec. 2015.

\bibitem{Scheirer_2013_TPAMI}
W.~J. Scheirer, A.~Rocha, A.~Sapkota, and T.~E. Boult.
\newblock Towards open set recognition.
\newblock {\em IEEE Transactions on Pattern Analysis and Machine Intelligence
  (T-PAMI)}, 36, July 2013.

\bibitem{Scholkopf:2001:ESH:1119748.1119749}
B.~Sch\"{o}lkopf, J.~C. Platt, J.~C. Shawe-Taylor, A.~J. Smola, and R.~C.
  Williamson.
\newblock Estimating the support of a high-dimensional distribution.
\newblock {\em Neural Comput.}, 13(7):1443--1471, 2001.

\bibitem{kextremes}
A.~Schultheiss, C.~K{\"{a}}ding, A.~Freytag, and J.~Denzler.
\newblock Finding the unknown: Novelty detection with extreme value signatures
  of deep neural activations.
\newblock In {\em Pattern Recognition - 39th German Conference, Proceedings},
  pages 226--238, 2017.

\bibitem{VGG}
K.~Simonyan and A.~Zisserman.
\newblock Very deep convolutional networks for large-scale image recognition.
\newblock {\em CoRR}, abs/1409.1556, 2014.

\bibitem{DBLP:journals/corr/SimonyanZ14a}
K.~Simonyan and A.~Zisserman.
\newblock Very deep convolutional networks for large-scale image recognition.
\newblock {\em CoRR}, 2014.

\bibitem{Tax:2004:SVD:960091.960109}
D.~M.~J. Tax and R.~P.~W. Duin.
\newblock Support vector data description.
\newblock {\em Mach. Learn.}, 54(1):45--66, 2004.

\bibitem{Turk1991}
M.~Turk and A.~Pentland.
\newblock Eigenfaces for recognition.
\newblock {\em Journal of Cognitive Neuroscience}, 3(1):71--86, 1991.

\bibitem{OOD}
A.~Vyas, N.~Jammalamadaka, X.~Zhu, D.~Das, B.~Kaul, and T.~L. Willke.
\newblock Out-of-distribution detection using an ensemble of self supervised
  leave-out classifiers.
\newblock In {\em Computer Vision - {ECCV} 2018 - 15th European Conference,
  Munich, Germany, September 8-14, 2018, Proceedings, Part {VIII}}, pages
  560--574, 2018.

\bibitem{Wright:2009:RFR:1495801.1496037}
J.~Wright, A.~Y. Yang, A.~Ganesh, S.~S. Sastry, and Y.~Ma.
\newblock Robust face recognition via sparse representation.
\newblock {\em IEEE Trans. Pattern Anal. Mach. Intell.}, 31(2):210--227, Feb.
  2009.

\bibitem{yosinski-2015-ICML-DL-understanding-neural-networks}
J.~Yosinski, J.~Clune, A.~Nguyen, T.~Fuchs, and H.~Lipson.
\newblock Understanding neural networks through deep visualization.
\newblock In {\em Deep Learning Workshop, International Conference on Machine
  Learning (ICML)}, 2015.

\bibitem{zhou2017places}
B.~Zhou, A.~Lapedriza, A.~Khosla, A.~Oliva, and A.~Torralba.
\newblock Places: A 10 million image database for scene recognition.
\newblock {\em IEEE Transactions on Pattern Analysis and Machine Intelligence},
  2017.

\end{thebibliography}
	}

\end{document}